# Performance Evaluation of Large Language Models in Bangla Consumer Health Query Summarization

Ajwad Abrar, Farzana Tabassum, Sabbir Ahmed
Department of Computer Science and Engineering, Islamic University of Technology, Gazipur, Bangladesh
Email: {ajwadabrar, farzana, sabbirahmed}@iut-dhaka.edu

*Abstract*—Consumer Health Queries (CHQs) in Bengali (Bangla), a low-resource language, often contain extraneous details, complicating efficient medical responses. This study investigates the zero-shot performance of nine advanced large language models (LLMs): GPT-3.5-Turbo, GPT-4, Claude-3.5-Sonnet, Llama3-70b-Instruct, Mixtral-8x22b-Instruct, Gemini-1.5-Pro, Qwen2-72b-Instruct, Gemma-2-27b, and Athene-70B, in summarizing Bangla CHQs. Using the BanglaCHQ-Summ dataset comprising 2,350 annotated query-summary pairs, we benchmarked these LLMs using ROUGE metrics against Bangla T5, a fine-tuned state-of-the-art model. Mixtral-8x22b-Instruct emerged as the top performing model in ROUGE-1 and ROUGE-L, while Bangla T5 excelled in ROUGE-2. The results demonstrate that zero-shot LLMs can rival fine-tuned models, achieving high-quality summaries even without task-specific training. This work underscores the potential of LLMs in addressing challenges in low-resource languages, providing scalable solutions for healthcare query summarization.

*Index Terms*—Large Language Models (LLMs), Bengali Consumer Health Queries, Abstractive Summarization, Zero-shot Learning, ROUGE Metrics Evaluation

## I. Introduction

The rapid expansion of online health consultation platforms has made Consumer Health Queries (CHQs) a prevalent form of medical inquiry. Patients frequently utilize these platforms to obtain advice, resulting in a considerable burden on healthcare professionals, who are required to respond manually to these inquiries [1]. A key challenge posed by CHQs is the frequent inclusion of irrelevant or extraneous details, which complicates the identification of critical information by medical professionals [2]. Abstractive text summarization involves creating a concise version of a text that preserves its essential information [3], presents a promising solution to this issue.

In recent years, several models tailored for Bengali, such as BanglaBERT [4] and BanglaT5 [5], have achieved impressive results across various Bengali NLP tasks [6, 7]. However, these models typically require fine-tuning on large, domain-specific annotated datasets, which are often unavailable for Bengali due to its underrepresentation in the field of NLP [8]. Large language models (LLMs) have shown remarkable proficiency in generating high-quality abstractive summaries, reducing the need for extensive fine-tuning and domain-specific data [9]. This makes them particularly beneficial for under-resourced languages like Bengali, where the availability of annotated datasets is limited.

While LLMs have demonstrated substantial success in English and other Indo-European languages [10, 11], their capabilities in Bengali, particularly for specialized tasks like CHQ summarization, remain underexplored. This study aims to fill this gap by evaluating the zero-shot performance of nine state-of-the-art large language models (LLMs): GPT-3.5-Turbo [12], GPT-4 [13], Claude-3.5-Sonnet [14], Llama3-70B-Instruct [15], Mixtral-8x22B-Instruct-v0.1 [16], Gemini-1.5-Pro [17], Qwen2-72B-Instruct [18], Gemma-2-27B [19], and Athene-70B [20] for the task of summarizing Bengali consumer health queries (CHQs). To the best of our knowledge, this is the first comprehensive study evaluating LLMs specifically for summarization tasks in the Bengali language within the healthcare domain.

## II. Literature Review

Text summarization has garnered significant research interest, leading to notable progress in both extractive and abstractive techniques [22, 23]. However, the task of summarizing CHQs has been inadequately addressed in existing literature. The limited research and resources available in languages other than English underscore a notable deficiency in the literature, particularly regarding the advantages of CHQ summarization for densely populated areas such as Bangladesh, where healthcare professionals frequently face overwhelming demands [24].

Previous studies have demonstrated the potential of large language models (LLMs) in CHQ summarization tasks. For instance, Jahan *et al.* conducted a comparative analysis between BioBART [25] and ChatGPT (GPT-3.5-turbo) in a zero-shot setting [10]. Their results indicated that ChatGPT outperformed BioBART, especially in cases where BioBART lacked domain-specific training data, underscoring ChatGPT's efficacy in low-resource environments. This emphasizes the capability of LLMs in summarization tasks, particularly in contexts where annotated datasets are limited, such as Bengali CHQs.

Table I: Sample Summaries from the BanglaCHQ-Summ Dataset [21]

| Original Question | Annotated Summary |
|---|---|
| আমার বয়স ৩০। অনেক দিন ধরে কোমরের ব্যথা করে। আমি ডাক্তার দেখিয়েছি। এক্সরে করার পর আমাকে নাপ্রক্সেন ১০ দিন ধরে খেতে বলে দিয়েছে। আর সাথে ক্যালসিয়াম, নিলটন খেতে দিয়েছে ১ মাস। কিন্তু এ পর্যন্ত ব্যথা যায় নাই। কোমরের প্রেসার দিলে ব্যথা করে। এখন কি করা যায়। | বয়স ৩০ । কোমরে ব্যথা । ডাক্তার এক্সরে করলে ন্যাপ্রক্সিন ১০ দিন ২ বেলা , ক্যালসিয়াম ও নিবসন ১ মাস খেতে দেয় । ব্যথা কমেনি । কি করণীয় ? |
| আমার আম্মুর পিঠের ঠিক মাঝে ভীষণ জ্বলে প্রায় এক বছর ধরে এমন হয়। ডাক্তার দেখানো হয়েছে শুধু গ্যাস্ট্রিকের ঔষধ দেয়। কিন্তু কোন কাজ হয়না। এখন খুব জ্বলে। মাঝে মাঝে একটু কম থাকে। কোন বিশেষজ্ঞ দেখালে ভাল হয়? আর কি সমস্যা হতে পারে? বিশেষজ্ঞদের পরামর্শ চাই। | পিঠের ঠিক মাঝে ভীষণ জ্বলে , ১ বছর ধরে । ডাক্তার গ্যাস্ট্রিকের ওষুধ দেয় , কোন কাজ হয়না । |
| আমার উচ্চতা ৫ . ৯ " ( ৬০ কেজি ) । মাস খানেক আগে আমার হালকা পাইলসের সমস্যা হয়েছিলো । নিয়মতান্ত্রিক খাওয়া দাওয়া আর হোমিও ওষধে কম এখন । আমি এখন জানতে চাচ্ছি এমতাবস্থায় কৃমিনাশক ট্যাবলেট খেতে পারবো কিনা এবং কৃমিনাশক ট্যাবলেট খাওয়ার নিয়ম কি ? আমি গত ৪ / ৫ মাসে কৃমিনাশক ট্যাবলেট খায়নি । | হালকা পাইলসের সমস্যা ছিল । এখন কৃমিনাশক ট্যাবলেট খাওয়া যাবে কি এবং খাওয়ার নিয়ম কি ? |

Further study has investigated the efficacy of LLMs in medical summarization tasks. Chen *et al.* [26] performed a systematic evaluation of open-source models, including Llama2 and Mistral, with GPT-4 acting as the assessor focusing on metrics such as coherence, fluency, and relevance. In parallel, Askari *et al.* [27] investigated paraphrasing tasks across various LLMs, including GPT-3.5-turbo, Llama2-13B, Mistral-7B, and Dolly-v2-7B, demonstrating the varying performance of models based on task complexity and size.

Jahan *et al.* [28] evaluated the performance of four leading large language models—GPT-3.5, PaLM-2, Claude-2, and Llama-2—across six benchmark biomedical tasks, including summarization, utilizing the MeQSum dataset for English CHQs. Although Claude-2 exhibited better performance relative to other LLMs, it still fell short against the state-of-the-art BioBART model when adequate in-domain training data was present.

Despite these advancements, there remains a lack of comprehensive evaluation of LLMs for Bengali CHQ summarization. This study addresses this gap by evaluating the zero-shot performance of nine state-of-the-art LLMs on Bengali CHQ summarization, offering insights into the capabilities of these models for under-resourced languages in the healthcare domain.

III. METHODOLOGY

This study evaluates the effectiveness of LLMs in summarizing Bangla consumer health queries in zero-shot settings. To achieve this, we utilized the BanglaCHQ-Summ dataset and evaluated the performance of nine LLMs for the assessment.

*A. Dataset*

The BanglaCHQ-Summ dataset, introduced in [21], is the first publicly available dataset specifically designed for consumer health query (CHQ) summarization in Bangla. It comprises 2,350 pairs of questions and their corresponding summaries. The data was sourced from a popular health forum[1] frequented by Bangla speakers, consisting of representative health queries collected from a platform that features user-submitted questions and certified medical responses. Given that the data was sourced from a public forum, it is reasonable to assume that the users possessed an average level of medical knowledge. The dataset covers 32 different health categories, ensuring a broad range of health topics. Sample consumer health queries along with their corresponding summaries are presented in Table I.

*B. Models*

Our selected nine LLMs for assessment can be categorized into proprietary and open-source models:

**Proprietary Models:**

- **GPT-3.5-Turbo**: The model was built on the Transformer architecture and consists of 175 billion parameters. It shows superior performance in large-scale NLP tasks. GPT-3.5-Turbo employs autoregressive decoding to generate coherent and fluent text, demonstrating high efficacy in tasks such as question-answering, summarization, and the comprehension of complex language patterns [29].
- **GPT-4**: It further improves the parameter efficiency and fine-tuning, leading to enhanced context understanding and refined text generation. It proficiently addresses multi-step reasoning tasks and delivers accurate outputs in summarization, translation, and creative text generation, demonstrating improved alignment for superior, human-like interactions [13].
- **Claude-3.5-Sonnet**: The emphasis is on producing nuanced and contextually aware responses for intricate inquiries. Utilizing around 52 billion parameters, it employs reinforcement learning from human feedback (RLHF) to enhance the alignment of its responses, thereby demonstrating proficiency in generating detailed and sensitive responses in complex conversational tasks [30].

---
[1] https://daktarbhai.com/

- **Gemini-1.5-Pro**: Developed by Google DeepMind, the model incorporates advanced transformer optimization methods and multitask learning functionalities. It utilizes reinforcement learning from external knowledge sources, enhancing its adaptability for tasks including text classification, natural language inference, and conversational AI [17].

**Free and Open-Source Models:**

- **Llama3-70b-Instruct**: The architecture comprises 70 billion parameters, enabling it to manage complex reasoning, sequence processing, and instruction-based tasks, with particular proficiency in multi-turn dialogue and intricate problem-solving. Llama3-70b enhances its transformer architecture for sophisticated NLP tasks, facilitating improved generalization across diverse applications [31].
- **Mixtral-8x22b-Instruct**: Utilizing eight groups of 22 billion parameters each facilitates effective parallelization. The multitasking capabilities render it appropriate for instruction-following tasks across diverse domains, while its computational efficiency ensures suitability for scalable tasks without compromising accuracy [16].
- **Qwen2-72b-Instruct**: It prioritizes the comprehension of user intent and the optimization of tasks specific to that intent. Featuring 72 billion parameters, this model is engineered to improve performance in context-rich and task-specific environments. It emphasizes natural language comprehension, enabling it to interpret nuanced inquiries and deliver contextually pertinent responses [18].
- **Gemma-2-27b**: Utilizing its 27 billion parameters to address complex and nuanced inquiries, its more compact architecture relative to larger models enables specialization in domain-specific sectors. Gemma-2-27b demonstrates significant efficacy in sectors including law, medicine, and technical disciplines, where accuracy and specialized knowledge are essential [19].
- **Athene-70B**: Athene-70B, optimized from Meta AI's Llama-3-70B model by Nexusflow, aims to improve the efficacy of the original architecture in practical applications. With 70 billion parameters, it has been optimized for enhanced generalization across various tasks, including conversational AI and advanced analytics [2].

### C. Prompt Design and Evaluation

We designed a zero-shot learning framework to benchmark the performance of these LLMs effectively. This approach allows us to evaluate how well these models generalize to domain-specific tasks without the benefit of fine-tuning, providing insights into their capabilities in summarizing Bangla consumer health queries.

Initially, we asked the LLMs to summarize the CHQs, but some models (e.g., Gemma-2-27b, Gemini 1.5 Pro, Llama-3-70B-Instruct) responded in English instead of Bangla. To correct this, we explicitly instructed them to generate responses in Bangla. Additionally, certain models (e.g., Claude-3.5-Sonnet, Athene-70B) provided overly detailed summaries, leading us to further refine the prompt to focus solely on the most essential information. After several iterations, the final optimized prompt was developed, ensuring concise summaries in Bangla. Finally, we used the following optimized prompt to obtain responses from the LLMs to generate concise summaries of the input queries:

> **Prompt:** "Provide a concise summary of the following long Bangla consumer health query. Focus on extracting only the critical and essential details, and ensure the output is a brief Bangla paragraph. Please avoid including any unnecessary information beyond the core query. Note: Kindly provide only the summarized output in Bangla.
> **[INPUT:]** "

### IV. RESULTS AND DISCUSSION

We present the performance evaluation of various LLMs for Bangla CHQ summarization, comparing their results against the current SOTA fine-tuned model, Bangla T5, as shown in Table II.

### A. Performance Evaluation

To evaluate the performance of the models on CHQ summarization, we used ROUGE-1 (R1), ROUGE-2 (R2), and ROUGE-L (RL) metrics [32]. The evaluation results indicate that LLMs like Mixtral-8x22B-Instruct-v0.1, GPT-4 and others, despite not being fine-tuned on the specific task of Bangla CHQ summarization, perform remarkably well compared to the state-of-the-art model, Bangla T5, which is a fine-tuned model on CHQs.

For instance, Mixtral-8x22B-Instruct-v0.1 outperforms Bangla T5 achieving the highest scores in ROUGE-1 (51.36) and ROUGE-L (49.17), while Bangla T5 leads in ROUGE-2 (29.11), a bi-gram overlap measure. Other models like GPT-4 and Llama-3-70B-Instruct also perform competitively across all metrics. However, various models exhibited suboptimal performance in this assessment. Athene-70B demonstrates poor performance across all metrics, recording an R1 score of 41.61 and the lowest ROUGE-L score of 40.14 among the assessed models. Similarly, Qwen2-72B-Instruct exhibits lower performance metrics, notably in ROUGE-1 (44.24) and ROUGE-L (42.06), indicating difficulties with content overlap and sequence fluency in this summarization task.

The relatively lower ROUGE-2 (R2) scores for zero-shot LLMs compared to Bangla T5 can be attributed to the nature of bi-gram evaluation. ROUGE-2 emphasizes the assessment of bi-gram overlaps, which are critical for evaluating a model's capacity to produce word sequences that maintain local context and coherence. Bangla T5, as a fine-tuned model for, has received task-specific training on the Bengali language, likely improving its capacity to generate bi-grams that align more closely with the ground truth summaries. In contrast, the LLMs, while possessing general language capabilities, do not exhibit fine-tuning for this particular domain. Their performance on R2 declines due to the lack of task-specific linguistic patterns and domain knowledge, resulting in less accurate bi-gram predictions and diminished local sequence consistency.

---
[2] https://nexusflow.ai/blogs/athene

**Table II:** Performance comparison between zero-shot LLMs and the fine-tuned state-of-the-art model on the BanglaCHQ-Summ dataset. Best results are highlighted in bold.

| Model | R1 | R2 | RL |
|---|---|---|---|
| GPT-3.5 Turbo | 45.17 | 10.56 | 44.33 |
| GPT-4 | 49.74 | 15.26 | 47.95 |
| Claude 3.5 Sonnet | 45.72 | 11.79 | 43.88 |
| Llama-3-70B-Instruct | 48.96 | 15.14 | 47.51 |
| Mixtral-8x22B-Instruct-v0.1 | **51.36** | 14.41 | **49.17** |
| Gemini 1.5 Pro | 48.92 | 12.72 | 46.62 |
| Qwen2-72B-Instruct | 44.24 | 11.08 | 42.06 |
| Gemma-2-27B | 46.56 | 12.27 | 44.17 |
| Athene-70B | 41.61 | 8.58 | 40.14 |
| Bangla T5 [21] | 50.05 | **29.11** | 48.35 |

**Table III:** Average word count of summarized text by different LLMs on the BanglaCHQ-Summ dataset.

| Model | Word Count |
|---|---|
| GPT-3.5 Turbo | 24 |
| GPT-4 | 37 |
| Claude 3.5 Sonnet | 36 |
| Llama-3-70B-Instruct | 19 |
| Mixtral-8x22B-Instruct-v0.1 | 33 |
| Gemini 1.5 Pro | 31 |
| Qwen2-72B-Instruct | 37 |
| Gemma-2-27B | 26 |
| Athene-70B | 23 |

### B. Analysis of Summary Length and Quality

We further observe that GPT-4 and Qwen2-72B-Instruct tend to generate significantly longer summaries (37 words on average) compared to the gold reference summaries (26 words on average). Interestingly, the best-performing model, Mixtral-8x22B-Instruct-v0.1, produces summaries with 33 words on average, slightly longer than the reference but still concise (Table III). Another notable observation is that Llama-3-70B-Instruct generates much shorter summaries, with an average of 19 words, while achieving performance nearly on par with GPT-4 in terms of ROUGE scores. This suggests that Llama-3-70B-Instruct is capable of delivering high-quality summaries with almost half the word count of GPT-4, highlighting its efficiency in text generation.

The evaluation of these LLMs in a zero-shot setting, without any task-specific fine-tuning, yet their close competition with a model fine-tuned specifically for Bangla, underscores their impressive generalization capabilities. This is particularly evident when examining the summaries generated by Bangla T5, Mixtral-8x22B-Instruct-v0.1, and GPT-4 in Table IV. The outputs reveal that these models produce remarkably similar summaries, effectively capturing essential medical information. This convergence in performance between fine-tuned models and LLMs, even in specialized domains such as bioinformatics, suggests that the gap is narrowing. These findings highlight the potential of large pre-trained models to excel in specialized tasks like Bangla query summarization, even in the absence of task-specific training.

### C. Impact of Pronunciation Variability

In analyzing the performance of various LLMs on Bangla text summarization, we observed that some summaries, particularly those generated by LLMs, exhibited lower ROUGE scores. Upon further investigation, we identified that pronunciation mismatches in Bangla were a contributing factor. The same word was often represented with different pronunciations in the reference (gold) summary and the LLM-generated summaries, leading to score discrepancies. This issue highlights a potential limitation in the models' handling of phonetic variations, which could explain the reduced scores in certain cases. These findings suggest that pronunciation variability in Bangla is a critical factor affecting LLM performance, warranting further research to better address these linguistic nuances and improve summarization accuracy.

In contrast, when we analyzed the top-performing models—Mixtral-8x22B-Instruct-v0.1, GPT-4, Llama-3-70B-Instruct, and Gemini 1.5 Pro—based on the R1 score, we observed a notable correlation between their performance and the word count of the generated summaries. Specifically, in instances where these models performed better collectively, the predicted summaries closely matched the word count of the gold standard summary. For example, in one case, the gold summary had a length of 24 words, while the average word count of the model-generated summaries was around 26. This consistency in word count appears to positively impact the models' performance.

## V. Conclusion

This study explores the potential of LLMs in the task of CHQ summarization, specifically in zero-shot settings. The results reveal that despite the lack of task-specific fine-tuning, LLMs demonstrate strong performance, coming close to and, in some cases, matching the performance of fine-tuned state-of-the-art models like Bangla T5. This underscores the growing capabilities of modern pre-trained models to generalize effectively across different tasks and languages. While this progress is promising, there is still a considerable journey ahead to fully bridge the gap between fine-tuned models and zero-shot LLMs, especially for specialized tasks in low-resource languages like Bangla. However, the decreasing performance gap suggests that the need for extensive fine-tuning may become less critical as LLM architectures continue to evolve. Future research will need to focus on improving the adaptability and efficiency of these models, ensuring even greater applicability in resource-constrained settings.

Table IV: Examples of summaries produced by benchmark models for comparison

| | Sample 1 |
|---|---|
| **Original Question** | আমার বয়স ৩০। অনেক দিন ধরে কোমরের ব্যথা করে। আমি ডাক্তার দেখিয়েছি। এক্সরে করার পর আমাকে ন্যাপ্রক্সেন ১০ দিন ধরে খেতে বলে দিয়েছে। আর সাথে ক্যালসিয়াম, নিলটন খেতে দিয়েছে ১ মাস। কিন্তু এ পর্যন্ত ব্যথা যায় নাই। কোমরের প্রেসার দিলে ব্যথা করে। এখন কি করা যায়। |
| **BanglaT5 Summary [21]** | বয়স ৩০। কোমরে ব্যথা। ডাক্তার ন্যাপ্রক্সিন ১০ দিন দু বেলা করে দেয়। সাথে ক্যালসিয়াম, নিবলন খেতে দেয় ১ মাস। কিন্তু ব্যথা যায় নাই। প্রেসার দিলে ব্যথা করে। কি করনীয়? |
| **Mixtral-8x22B-Instruct-v0.1 Summary** | আমি ৩০ বছর বয়সী এবং কোমর ব্যথা করছি । ডাঃ এক্সরে করেছেন এবং ন্যাপ্রক্সিন, ক্যালসিয়াম ও নিবলন প্রদান করেছেন । কিন্তু ব্যথা যায় নাই । কোমরে প্রেসার দিলে ব্যথা করে । এখন কি করবো? |
| **GPT-4 Summary** | আমার বয়স ৩০ বছর। অনেকদিন ধরে কোমরে ব্যথা হচ্ছে, ডাক্তারের পরামর্শে ন্যাপ্রক্সিন, ক্যালসিয়াম ও নিবলন খেয়েছি, কিন্তু ব্যথা কমেনি। কোমরে চাপ দিলে ব্যথা বাড়ে। এখন আমি কী করতে পারি? |
| | Sample 2 |
| **Original Question** | আমার আম্মুর পিঠের ঠিক মাঝে ভীষণ জ্বলে প্রায় এক বসর ধরে এমন হয়। ডাক্তার দেখানো হয়েছে শুধু গ্যাস্ট্রিকের ওষুধ দেয়। কিন্তু কোন কাজ হয়না। এখন খুব জ্বলে । মাঝে মাঝে একটু কম থাকে । কোন বিশেষজ্ঞ দেখালে ভাল হয়? আর কি সমস্যা হতে পারে? বিশেষজ্ঞদের পরামর্শ চাই। |
| **BanglaT5 Summary [21]** | পিঠের ঠিক মাঝে ভীষণ জ্বলে প্রায় এক বছর ধরে । ডাক্তার গ্যাস্ট্রিকের ওষুধ দেয়। কাজ হয়না। এখন খুব জ্বলে । মাঝে মাঝে একটু কম থাকে। কোন বিশেষজ্ঞ দেখালে ভালো হবে? আর কি সমস্যা হতে পারে? |
| **Mixtral-8x22B-Instruct-v0.1 Summary** | আমার পিঠে জ্বলে এক বসর ধরে আছে, ডাক্তারের দেওয়া গ্যাস্ট্রিক ওষুধ কাজ করেনি। এখন এটা খুব বেড়ে গেছে, কোন বিশেষজ্ঞের পরামর্শ চাই। |
| **GPT-4 Summary** | আমার আম্মুর পিঠের মাঝখানে এক বছর ধরে ভীষণ জ্বালা হচ্ছে, ডাক্তার গ্যাস্ট্রিকের ওষুধ দিলেও কাজ হয়নি। এখন জ্বালাটা আরও বেড়ে গেছে। বিশেষজ্ঞের পরামর্শ প্রয়োজন এবং অন্য কোন সমস্যা হতে পারে কিনা জানতে চাই। |